**PointsToWood: A deep learning framework for complete canopy leaf-wood segmentation of TLS data across diverse European forests.**

Owen, H. J. F.[1,*], Allen, M. J. A.[1], Grieve S.W.D.[2], Wilkes P.[3], Lines, E. R.[1]

[1]Department of Geography, University of Cambridge, Downing Site, Cambridge.

[2]School of Geography, Queen Mary University of London, London.

[3]Royal Botanical Gardens, Kew, Wakehurst Place, Sussex.

*corresponding author, ho304@cam.ac.uk

**Abstract**

1. Point clouds from Terrestrial Laser Scanning (TLS) are an increasingly popular source of data for studying plant structure and function, but typically require extensive manual processing to extract ecologically important information. One key task is the accurate semantic segmentation of different plant material within point clouds, particularly wood and leaves, which is required to understand plant productivity, architecture, competition, space optimisation and physiology, and is a key step in common approaches to individual tree extraction. Existing automated semantic segmentation methods are primarily developed for single ecosystem types, and whilst they show good accuracy for biomass assessment from the trunk and large branches, often perform less well within the crown.

2. In this study, we demonstrate a new framework that uses a deep learning architecture newly developed from PointNet++ and pointNEXT for processing 3D point clouds to provide a reliable semantic segmentation of wood and leaf in TLS point clouds from the tree base to branch tips, trained on data from diverse mature European forests. Our model uses meticulously labelled data combined with voxel-based sampling, neighbourhood rescaling, and a novel gated reflectance integration module embedded throughout the feature extraction layers. We evaluate its performance across open datasets from boreal, temperate, Mediterannean and tropical regions, encompassing diverse ecosystem types and sensor characteristics.
3. Our results show consistent outperformance against the most widely used PointNet++-based approach for leaf/wood segmentation on our high-density TLS dataset collected across diverse mixed forest plots across all major biomes in Europe. We also find consistently strong performance tested on others' open data from China, Eastern Cameroon, Germany and Finland, collected using both time-of-flight and phase-shift sensors, showcasing the transferability of our model to a wide range of ecosystems and sensors. We develop a new evaluation metric to weight performance in the outer parts of the canopy, such as in twigs and small branches, and find our model consistently outperform the most widely used PointNet++-based approach.
4. To promote transparency, reproducibility, and advancements, we've openly shared all labelled data, code, and model weights, establishing a comprehensive benchmark for future evaluations.

**Introduction**

Terrestrial Laser Scanning (TLS) is revolutionising how we monitor forest ecosystems. TLS scanners produce accurate highly detailed three-dimensional point clouds which can be used to characterise the aboveground structure of forests, including foliage arrangement, woody biomass and habitat space for other dwelling organisms (Malhi et al., 2018; Disney et al., 2019; Calders et al., 2020; Lines et al. 2022a). The detail in TLS data has led to widespread interest and their uptake by research groups, foresters and land managers, leading to the generation of massive amounts of data on forests globally (Global TLS: www.global-tls.net/). However, processing three-dimensional data to allow the extraction of meaningful metrics on forest structure and function remains very labour intensive and impractical at scale (Disney et al., 2018), meaning the full potential of these data has not yet been fully realised.

The biggest hurdle currently faced by scientists using these data is the separation of a raw point cloud into biophysically interpretable components, including the semantic segmentation of leaves, wood and other forest material, and the related, and often downstream, task of instance segmentation of individual trees. Semantically labelled data contain significant biological and ecological information that could be exploited in ecological studies. For example, quantifying the spatial arrangement and relative amounts of leaves and wood within a tree could inform understanding of biophysical properties of forests including energy and water fluxes and light use (Béland et al., 2014). Wood-only point clouds (not directly measurable in evergreen forests) allow direct estimation of aboveground timber volume and carbon storage (Burt et al., 2019) and could improve estimates of woody respiration (Malhi et al., 2018). Leaf-only point clouds allow estimates of leaf area (Zhu et al., 2023), and leaf size and angle (Stovall et al., 2021), which are important determinants of primary productivity and light use. Instance segmentation isolates individual trees, the fundamental unit of forest ecology, creating the opportunity for wholly new understanding of ecological processes including tree-tree interactions and forest dynamics (Owen et al. 2021; Lines et al. 2022a; Malhi et al. 2018).

Deep Learning (DL) tools have shown promise towards the goal of universal automated processing for TLS data. While significant progress has been made, fully automated instance segmentation that seamlessly handles complex forest structures remains a challenge. The use of semantic segmentation to separate wood clouds prior to individual tree detection is a promising approach, but current operational applications are primarily limited to relatively simple forest stands such as production monocultures, with clear separation between trees (e.g., Krisanski et al., 2021a). More diverse and intricate forest environments still present considerable hurdles for end-to-end solutions. This is in part because of the difficulty in obtaining enough high-quality data with sufficient fine scale information at crown boundaries to train large models, but also because generic DL frameworks require specific adaptation to the difficulties of working with 3D scanning data in forests. Such data have common and challenging characteristics in their data structure, including noise, occlusion and irregularity, in contrast with common benchmark datasets (such as semanticKITTI; Behley et al., 2019) used to develop and test models, which are mostly composed of simplified objects or street scenes. Further, TLS scanning data are very memory-hungry because they are both very dense (e.g. >100,000 points per m$^3$) and very detailed. This, along with the high structural variability within and between forests of different biomes and structures, and variety of sensors in use, means that methodological approaches developed for either specific tasks or ecosystems don't necessarily transfer to different contexts (Lines et al. 2022b).

A number of statistical and rules-based approaches to semantic and instance segmentation for TLS forest data have been proposed, but these have limitations. These include requiring individual tree point clouds for leaf/wood separation (TLSeparation; Vicari et al., 2023), requiring predefined feature engineering for semantic segmentation (Wan et al., 2021), or requiring context-dependent allometric assumptions of tree structure for tree separation (Burt et al., 2019; Tao et al 2015). DL-based approaches have been able to bypass some of these issues, but developments have been based on single or simple ecosystem datasets (Krisanski et al., 2021a; Wan et al., 2021; Wang et al., 2021; Morel et al., 2020; Jiang et al., 2023; Bai et al., 2023), and their transferability to other systems or breadth of application is typically not tested. Indeed, it is not clear whether a single model may be suitable for all forest types or sensors, or if ecosystem and sensor-specific approaches are needed.

The most significant early development in DL semantic segmentation in forests is the Forest Structural Complexity Tool (FSCT) from Krisanski et al. (2021b). This was one of the first approaches of semantic segmentation of forest point clouds using DL and was trained on ground and UAV-based LiDAR data from *Pinus radiata* and *Eucalyptus* monocultures in Australia. FSCT's advantages over non-DL methods include its ability to semantically segment blocks of data (rather than requiring individual trees), its sensor-agnostic design, and its downstream tree segmentation module. Published as an open tool, FSCT has contributed to several further advancements in the field, including the hybrid semantic-instance segmentation software, TLS2Trees (Wilkes et al., 2023). This multi-biome, TLS-specific tool applies the FSCT semantic module to perform DL semantic segmentation on blocks of data, isolating wood points, and using these to identify individual tree skeletons, statistically reattaching leaf points after instance segmentation. This relies upon accurate semantic labels, but FSCT has difficulty classifying fine-scale features, typically performing best on large woody elements. Whilst the open nature of FSCT permits retraining, labels, particularly semantic labels, are extremely labour-intensive to generate, and few high-quality open labelled data are currently available.

Here, we develop a new semantic leaf-wood segmentation approach specifically tailored to perform well on both small- and large-scale woody structural features of trees. We take an existing deep learning framework as our starting point and make both computational and structural modifications to improve accuracy and minimise computational overhead at high resolutions, whilst increasing transferability to multiple forest ecosystem types by training on multi-biome data. We focus development on the most challenging aspect of the semantic task - segmentation of small woody features that are often poorly resolved in data (due to sensor characteristics and occlusion) but which are crucial for accurate crown structure and crown-crown separation. To do this, we create a binary leaf-wood model with wood as the target (positive) class. The model is trained on a database of TLS point clouds collected from diverse and structurally varying boreal, temperate and Mediterranean forests within the FUNDIV ([www.fundiveurope.eu](www.fundiveurope.eu)) network in Europe, which we publish openly. We train both biome-specific and pan-European models to establish whether ecosystem specific models offer any performance gains, or whether more generic models applicable to multiple forest types are feasible.

We compare our model results against the performance of the standard and widely used FSCT, assessing using both standard DL performance metrics and performance on small-scale woody elements of specific interest to ecologists. Our overarching aim is to develop an algorithm specifically targeted at high-resolution TLS data, distinguishing it from FSCT's sensor-agnostic approach. We compare our algorithm against both an off-the-shelf version of FSCT and an iteration retrained on our dataset. To do so, we use balanced accuracy, precision and recall. Furthermore, to assess performance on the most challenging outer peripheries of tree crowns, we developed and use a new metric: balanced accuracy weighted by path length (a measure of distance between points at the trunk base to points at branch tips), which emphasises performance at small scale canopy elements closer to branch tips. We evaluate our model's performance against both the off-the-shelf FSCT and a retrained FSCT models using our own data. We evaluate on unseen data from our forest plots, and four additional independent and open datasets collected by others, assessing performance on unseen ecosystem types and sensors. We present the code as an open-source software and provide both one pan-European model and three biome-specific variants to enable users to choose the most appropriate model for their application.

**Materials and Methods**

*Summary of approach*

In this study, we developed a new DL method that directly addresses a key task for ecological applications of TLS data from forests: semantic segmentation of leaves and wood throughout the canopy. Using xyz and reflectance data collected with a single, commonly used, sensor - a RIEGL VZ400i time-of-flight TLS - we present a redesign of past implementations of PointNet++ in this context (Qi et al., 2017; Krisanski et al., 2021a) to:

1) integrate weighted sampling using reflectance as a prior estimate of class to address severe class imbalance (majority leaf points) within the canopy;
2) apply a focal loss function to force the model to learn from small numbers of hard to classify points;

3) normalise neighbourhood coordinates by the neighbourhood size to improve network optimisation (following Qian et al., 2022);

4) improve downsampling to retain features of interest (small woody features),

5) Dynamic gating of reflectance during training and inference, enabling the model to selectively utilize reflectance data when beneficial and

6) implement fast and effective voxelisation during preprocessing.

Our approach begins with an efficient preprocessing pipeline (Figure 1), followed by a framework that uses Pointnet++ (Qi et al., 2017) as a backbone, with additional complexity from pointNEXT (Qian et al., 2022) to best capture fine scale details in a computationally efficient manner.

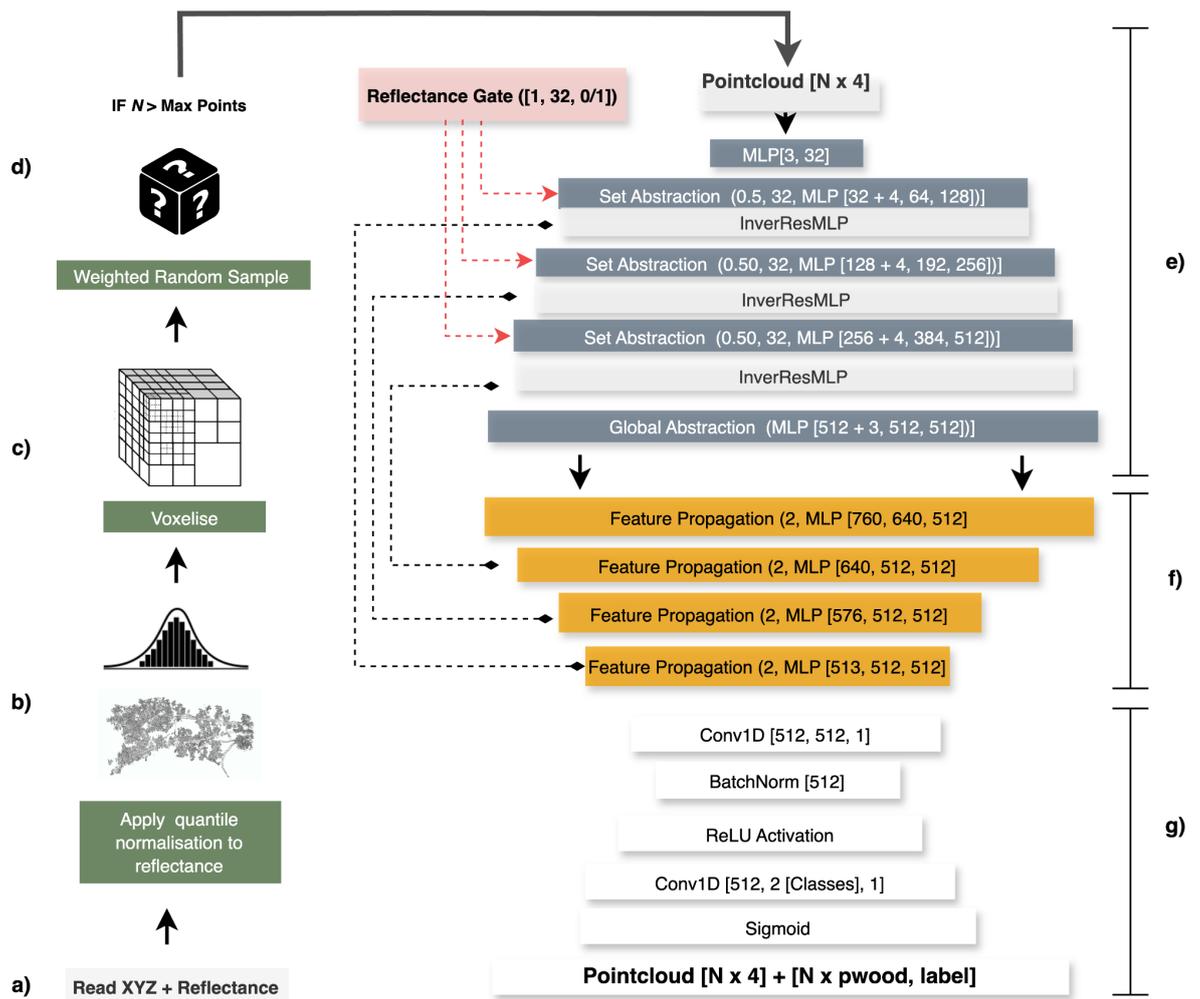

**Figure 1:** *Schematic of preprocessing (a-d) and model framework (e-g).*

*Data collection*

The TLS data in this study was collected from a subset of the FunDivEUROPE permanent forest plot network (www.fundiveurope.eu), established in 2011 for the specific purpose of testing the effects of biodiversity on ecosystem functioning in diverse mature forests across Europe. More details on the network sampling design and protocols are available in Baeten et al., (2013). We selected plots in three countries within the network to represent the three major forest biomes of Europe, located in Alto Tajo Natural Park in central Spain (Mediterranean), Białowieża forest in Eastern Poland (temperate), and forests around Joensuu in Eastern Finland (boreal). Plots include locally dominant tree species and range in species richness from 1 to 3 in Finland, 1 to 4 in Spain and 1 to 5 in Poland. The species mix includes evergreen conifers, deciduous broadleaf and evergreen broadleaf individuals, with the climatic range and disturbance history creating extensive variability in structure (see Ratcliffe et al., 2017). Within each country, we scanned a subset of the 30 m x 30 m FUNDIV plots using a Riegl VZ400i TLS instrument (RIEGL Gmbh, Horn, Austria), scanning at 600MHz and with an angular resolution of 0.04 mrad. All plots were scanned following a 10 m grid system with a minimum of 16 upright and 16 tilt scans (following Wilkes et al. 2017), with additional scans to minimise occlusion in dense areas, and outside the plot perimeter to capture edge trees. To ensure high-quality data with minimal noise, scanning was paused when wind conditions rose above 5 m/s (measured with an anemometer on the ground) or when gusts were visually evident.

Scans were co-registered in RiScan Pro and the rotation matrices along with the scan data were further processed using a bespoke pipeline that filters points with a deviation (a unitless measure of noise with Riegl instruments) above 15 and reflectance below -20 (both unitless metrics generated by the Riegl scanner), which we adapted from Wilkes et al. (2024). To create the train-validation dataset, we chose three plots from each country representing contrasting structure and diversity, and from each of these generated 10 m x 10 m data blocks (full plot height blocks), centred around the midpoint of plots, making nine blocks in total. We used two blocks from each country for training and retained one for validation. Similarly, we generated three 10 m x 10 m data blocks, for each country for a comprehensive independent assessment of model performance (see 'model evaluation' section below, which also includes details of additional open data collected by others used for evaluation).

*Manual data labelling*

To create our model, we first needed to label leaf and wood points in our data, for which we used a semi-automated approach informed by existing approaches followed by significant manual cleaning. Vicari et al. (2019) found anisotropy, verticality and linearity to be informative features for leaf-wood separation, so we created these geometric features at spatial scales of approx. 5 cm - 0.5 m (using CloudCompare, 2023). Alongside these, we used reflectance and xyz information for each point and labelled leaf-wood by thresholding these features. We followed this with intensive manual checking and cleaning to ensure high label quality, especially in smaller branches and twigs. Our dense scanning and labelling strategy mean that this dataset is an ideal candidate for training and testing the capabilities of processing algorithms, and it is published alongside this paper as a benchmark dataset for use by others (doi: 10.5281/zenodo.13268500).

*Downsampling and voxelization*

Typically, DL models require voxelised 3D data to be downsampled in order to carry out classification on point clouds, but done poorly this can result in the loss of important features, particularly small details. Here we define our target class as wood, and we performed a number of steps (Figure 1 a-e) prior to model initialisation to reduce target information loss in hard to classify regions of the point cloud. Previous DL forest semantic segmentation approaches have downsampled significantly, for example, FCST retains only 20k points per 6 x 6 x 8 m voxel (for comparison, the same sized block of our test-validation data in raw form can contain over 2 million points), likely leading to important information loss where branch sizes are smaller.

Our preprocessing routine creates voxels at two scales (2 m and 4 m, equating to 4 and 16 $m^3$ voxels) as input to the model, as follows. First, ground points are removed using the very efficient cloth simulation filter (Zhang et al., 2016), and then reflectance is normalised using quantile normalisation and normalised to within -1 and 1 (Figure 1b to make it comparable across our dataset. Then, we leverage the voxel grid function within torch geometric (Fey et al., 2019) to voxelise the point cloud at both resolutions using the GPU (Figure 1c). Next, each voxel is downsampled using a reflectance weighted random sampling approach if total points exceed the maximum limit of 16,384 points (Figure 1d. If no reflectance is present, random downsampling is used. The final output voxels (model input) consisted of x, y, z, and reflectance and were written to disk, with reflectance filled with zeros if undetected within the input. A comparison of output of a random downsampling versus our weighted strategy on the same sample of data can be seen in Figure 2.

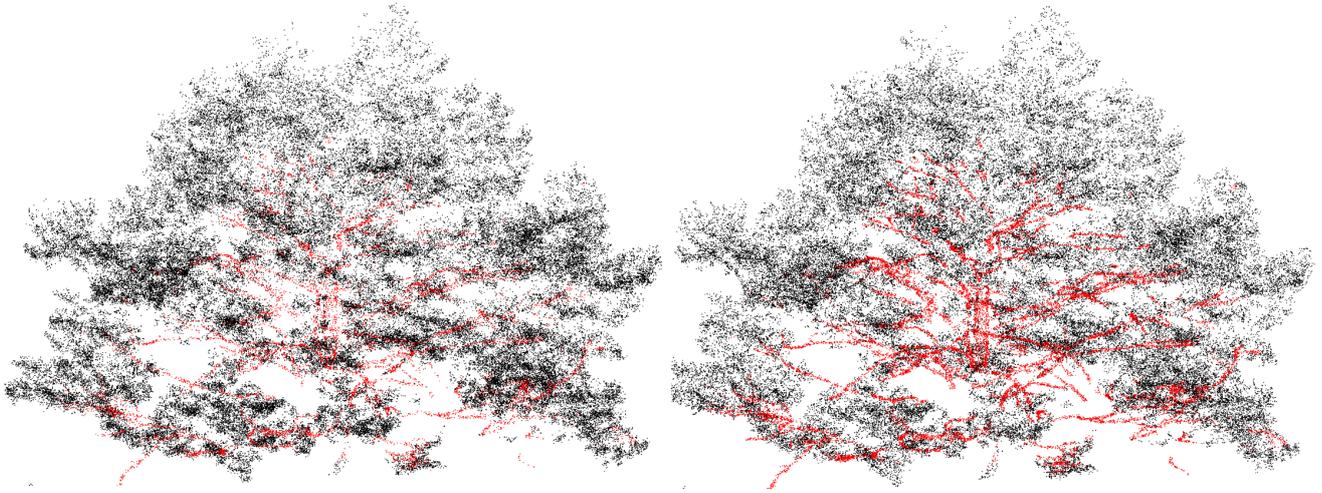

**Figure 2**: *Comparison of output between (left) a random and (right) our weighted downsampling strategy. Both clouds downsampled data to 100k points, and images were created with the same colour scales (red is wood, black is leaf, with darker tones showing more dense regions). The random downsample resulted in 89,824 leaf and 10,176 wood points, whereas our weighted downsample retained more than twice the number of wood points (77,871 leaf and 22,129 wood). Tree crown is approximately 8 metres wide.*

*Network architecture*

We developed a new classification network by modifying the architectures of PointNet++ and pointNEXT (Qian et al., 2022; Qi et al., 2017, Figure 1 f-h) and using FSCT as a starting point. Central to these frameworks are the set abstraction (SA) modules, which efficiently generate local features of reduced dimensionality by employing techniques including point sampling, nearest neighbours, and multilayer perceptrons, all nested within a structured hierarchy (see Figure 1f). The final encoding stage includes a global max pool, similar to FSCT, to extract the maximum feature from the preceding hierarchical layers. These local features are then combined and refined in a hierarchical fashion within feature propagation modules (FP), capturing information across all scales produced by the SA modules (Figure 1g). On input, each sample is individually normalised by dividing each point by the maximum distance, reducing the effects of varying scale across samples and datasets. In addition to sample normalisation, each focal point neighbourhood is also centred and rescaled by the maximum distance. This process reduces magnitudinal differences between neighbourhoods and set abstraction stages, thereby enhancing optimization. We modify the set abstraction modules to integrate reflectance information effectively, while also enabling learning in its absence. The model incorporates a learnable reflectance gating mechanism at each set abstraction stage, consisting of two linear layers that process reflectance data sample-by-sample within each batch. The output is passed through a Gumbel-Softmax operation, producing a binary decision for each sample. This approach enables the model to adaptively modulate the influence of reflectance data during the convolution process, allowing for dynamic, sample-specific integration with spatial coordinates. The use of Gumbel-Softmax ensures differentiability during training while maintaining discrete binary decisions during inference. This method effectively adjusts to scenarios where reflectance data may be absent, noisy, or unreliable, ensuring robust performance under varying conditions. To enhance the detection of fine-scale woody structures, we employ smaller neighbourhoods than FSCT and replace traditional furthest point sampling with voxel-based sampling. This type of sampling partitions the point cloud into a regular grid of volumetric pixels (voxels), selecting representative points from each occupied voxel to reduce data density while preserving spatial distribution. Additionally, inspired by Point-BERT (Yu et al., 2022), we replace voxel grid sampling with random sampling during training to enhance model robustness. In the first set abstraction stage, neighbourhood searches are constrained within a specific radius, while in subsequent stages, all 32 neighbours are gathered regardless of distance. This method preserves spatial integrity, ensures transferability across datasets with varying characteristics, and improves computational efficiency.

We apply model scaling strategies taken from pointNEXT (Qian et al., 2022) and Mobilenetv2 (Sandler et al., 2018) which include inverted residual bottleneck blocks within each SA step (Figure 1f). These blocks incorporate skip connections, improving gradient flow through the network by combining features from different levels of abstraction. Additionally, we leverage separate pointwise and depthwise convolutions to accelerate computations and enhance the efficiency of the model. This design enriches the feature sets by expanding the number of channels by a factor of four without the conventional computational overhead. Lastly, we apply batch normalisation throughout all except the first input layer, apply RELU activation within the internal structure, and a sigmoid for the final binary output. To ensure that our point clouds are centred for consistent processing, we centre the data by subtracting the mean.

*Loss function: prioritising the target wood class*

A striking characteristic of TLS forest point clouds is that many regions within them contain either almost entirely wood points (i.e. trunks) or almost entirely leaf points (leaves, grasses). But in the mixed regions - tree crowns - our target class, wood, is often outnumbered within the crown by the non-target class, leaves, by at least a factor of 10 (e.g. Figure 2). Further, woody elements scale in size by orders of magnitude from trunk to branch tips, whereas leaves usually do not vary in size substantially within a crown. This means our target class (wood) changes throughout the scene from a focal context perspective, while our non-target class experiences much less variation. On the edges of crowns where leaves are most prevalent, woody features become very small and heavily interlaced with leaf points. The challenge is further exacerbated by systematic properties of the TLS data, with increased beam width and occlusion with distance from the scanner resulting in reduced signal-to-noise ratio in the most challenging areas of the upper canopy. To ensure optimal performance in downstream tasks for these regions, the model must prioritize minimizing loss for wood points, even when they comprise only a small fraction of the data. To achieve this, we implemented a focal loss function and incorporated sample class weights to strike a balance between precision and accuracy. Specifically, we use the focal loss function proposed by Lin et al. (2017), setting the gamma parameter to 2.0 but alpha to None. This choice focuses the model's attention on points that are difficult to classify, while downweighting those for which the model is already confident. We also applied label smoothing as a known effective regularisation technique, with an alpha value of 0.1 (Szegedy et al., 2016).

*Data augmentation*

We applied data augmentation on the fly during training, as voxels are sampled from the disc during training time. To mitigate over-reliance on reflectance information and enhance model transferability to diverse datasets, we implemented two data augmentation techniques post-normalisation: (1) random replacement of 25% of reflectance values with the mean (zero, due to normalisation), following Qian et al. (2022), and (2) application of Gaussian noise ($\mu = 0$, $\sigma = 1.0$) to an independent 25% of reflectance data. This approach addresses the variability in reflectance availability across different conditions and sensors, and aims to guide the model to emphasise learning from the spatial dimensions (x, y, z), thereby potentially improving accuracy for other sensors. We also applied random 3D rotations (x, y, z) on 25% of samples. As the recent pointNEXT model (Qian et al., 2022) found negative effects on performance, we chose not to add noise as an additional augmentation strategy to the positional data.

*Model training and inference*

At train time, we used the AdamW approach (Loshchilov and Hutter, 2017) to optimise the model over 300 epochs with a batch size of 10. The learning rate was applied using a one cycle cosine decay schedule, with 10 warm up steps and a max learning rate of 0.001. Both the balanced accuracy and F1 score were calculated after each epoch, with the best model on the validation set saved during the training process. We trained four models: one pan-European model trained on all training blocks (six in total, two from each country), and three biome-specific models for boreal, temperate and Mediterranean forests separately (trained on two blocks of data from the relevant biome). The remaining blocks of data were retained for validation.

Our model makes predictions across overlapping data that must be amalgamated to produce a single probability and label for every data point in the input. The model generates predictions for overlapping regions, necessitating a consolidation process to derive a single probability and label per input data point. This consolidation is achieved through a k-nearest neighbour approach (k = 32) coupled with a weighted voting mechanism. For each point, class votes are accumulated based on thresholded prediction probabilities (threshold = 0.5 for wood classification), with the final label determined by the class with the highest vote tally. The associated probability is computed as the mean of all probabilities within the defined neighbourhood. Both training and inference were conducted on Ubuntu 20.04 LTS using a 32 core CPU with 256GB memory, and a Nvidia Quadro RTX 6000 GPU with 24GB video memory.

To ensure a fair and informative comparison with FSCT, we evaluated our model against both a retrained version using our dataset and the original off-the-shelf variant. The FSCT model was retrained on our dataset following the procedures described in Krisanski et al. (2021a): training for 300 epochs with a batch size of 8, and a learning rate decreasing from $5 \times 10^{-5}$ to $2.5 \times 10^{-5}$ after 150 epochs.

*Model evaluation with traditional and new context-specific metrics*

In order to evaluate our model's performance, we tested it against both our own data (unseen data chunks from each country/biome, labelled in the same way as the test-train dataset) and against high quality open data identified following an extensive search (see next section). We tested our three biome-specific models and our pan-European model against each other and against a version of FSCT retrained on our data (Table 1), as well as against the off-the-shelf version (Table S1). We evaluated model performance using traditional DL metrics (balanced accuracy, precision and recall), but also incorporated a new metric, path-length weighted accuracy, which weights the model performance at the crown peripheries (highest path length) higher, therefore showing model performance at the crucial crown-crown boundaries. Calculating path-length for each point requires the data to be segmented into individual trees. To segment individual trees in our data, we first ran TLS2trees (Wilkes et al., 2023) and then manually refined the output using tools within CloudCompare (CloudCompare, 2024). To derive path length values for each point, we used the shortest path functionality within the TLSeparation python package (Vicari et al., 2019) where a path is traced from the lowest coordinate (base of a tree) throughout the point cloud using the Dijkstra shortest path approach. The output is a distance value for every point calculated as the shortest path through the point cloud to the tree base.

*Model evaluation with third party data*

To assess our model's performance on diverse Terrestrial Laser Scanning (TLS) data, we used open-access datasets with high-quality, visually verified leaf/wood labels from varied ecosystems and sensors. These datasets, sourced from Wang et al. (2021), Wan et al. (2021), Weisser et al. (2024) and Mspace Lab. (2024), encompass a range of forest types and geographic locations. Wan et al. (2021) collected data from temperate forests in China, featuring two monoculture plots of Dahurian larch and Chinese poplar, collected using a Riegl VZ-1000 scanner. Wang et al. (2021) collected data from a semi-deciduous tropical forest in Eastern Cameroon, captured with a Leica C10 Scanstation. Weisser et al., (2024) includes a mixture of European deciduous (beech, maple, oak) and evergreen conifers (Scots Pine, Norway Spruce) collected using a Reigl VZ-400 scanner in Germany. Mspace Lab, (2024) collected data from an evergreen conifer plot in Finland, containing Norway spruce, silver birch, and Scots pine, scanned with a Leica HDS6200. The Finnish dataset included spatial (xyz) coordinates and intensity information, while the German dataset contained spatial (xyz) coordinates and reflectance data. The Chinese and Cameroonian datasets contained only spatial (xyz) coordinates. To standardise input for our model, we augmented datasets lacking reflectance information with a column of zeros, equivalent to mean reflectance in our normalisation approach. As the third-party datasets we used lack individually segmented trees, we did not generate balanced accuracy weighted by path length (BAP) for their outputs. This diverse dataset collection enabled an unparalleled robust evaluation of our model's adaptability to varying forest structures, sensor types, and data characteristics.

**Results**

We trained four models: three biome-specific models using our Spanish (Mediterranean), Polish (temperate) and Finnish (boreal) data separately, and one pan-European model using data from all three countries combined. We compared our model results with those from our retrained FSCT as a comparative baseline and found that all outperformed FSCT by most performance metrics, both for our dataset and others' (Table 1). For completeness we also present results of our model compared to the off-the-shelf FSCT in Table S1, where we have found our model performed better on all datasets, but we note that these are likely driven by differences in input data. For the third-party datasets, the re-trained FCST performed worse on all datasets than its off-the-shelf variant for balanced accuracy and recall, and for precision for three of the four datasets (compare Tables 1 and S1).

**Table 1** *Model performance metrics (0 to 1) for a) our pan-European model and b) our biome-specific models, both compared to FSCT outputs. The datasets in this study come from various sources with differing characteristics. Data from Finland, Spain, and Poland were collected by the authors and contain reflectance values. Additional datasets from another site in Finland, Germany, China, and Cameroon were obtained from publicly available open repositories. Among these public datasets, only those from Finland and Germany include reflectance information, while the China and Cameroon datasets contain only xyz coordinates. For the latter, reflectance values were assigned as 0 (mean reflectance, indicated by \*). Balanced accuracy, precision, recall, and our new metric path length weighted accuracy were used to assess model performance, with our model results given first in **bold** and FSCT results given second, in standard font. TOF is an abbreviation for time-of-flight.*

| a) **Pan European model** vs FSCT re-trained on our data | | | | |
|---|---|---|---|---|
| Country | Balanced accuracy (BA) | Precision | Recall | Balanced accuracy weighted by pathlength (BAP) |
| *Our data (TOF with reflectance)* | | | | |
| Finland | **.904**, .854 | **.943**, .788 | **.819**, .756 | **.883**, .811 |
| Poland | **.971**, .884 | **.865**, .913 | **.954**, .774 | **.960**, .848 |
| Spain | **.934**, .763 | **.910**, .946 | **.905**, .534 | **.905**, .726 |
| *Third party data* | | | | |
| Cameroon (TOF no reflectance) | **.952**, .840 | **.919**, .931 | **.921**, .691 | n/a |
| China (TOF no reflectance) | **.912**, .814 | **.673**, .781 | **.878**, .646 | n/a |
| Germany (TOF w/reflectance) | **.929**, .800 | **.893**, .939 | **.898**, .614 | n/a |
| Finland (phase-shift w/intensity) | **.825**, .793 | **.489**, .630 | **.890**, .689 | n/a |
| b) **Biome-specific models** vs FSCT re-trained on our data for each biome | | | | |
| Finland (boreal) | **.837**, .854 | **.976**, .788 | **.678**, .757 | **.808**, .811 |
| Poland (temperate) | **.941**, .884 | **.931**, .913 | **.888**, .774 | **.931**, .848 |
| Spain (Mediterranean) | **.935**, .763 | **.903**, .946 | **.910**, .534 | **.907**, .726 |

Our models had consistently higher balanced accuracy (BA), balanced weighted accuracy by pathlength (BAP) and recall than FCST, but performance evaluated with precision was more varied between the two approaches for both our pan-European and biome-specific models. Our model also showed improved predictive capacity on the unseen ecosystem datasets from Cameroon, China, Germany and Finland, despite two of these containing no reflectance information, suggesting transferability of our approach. For our data, all our models' BAP were consistently higher than those of the re-trained FSCT, demonstrating that our model has achieved our aim of improved classification ability in hard to classify peripheries of tree crowns. An example comparison between predictions of our pan-European model and FCST can be seen in Figure 3, showing the clear visual improvement in within-crown small woody detection characteristic of our results.

*Pan European vs biome-specific model performance*

Our pan-European model performed as well or better than each biome-specific model for both BA and BAP, and recall (Table 1) with the largest difference seen within our own Finland dataset (a decrease in BA of 6.7%). Biome-specific models outperformed the pan European model for precision in the boreal and temperate forests, but not the Mediterranean.

*Performance of our pan-European model and FSCT on our data*

Our model had consistently higher prediction accuracies for both BA and BAP than the retrained FSCT (Table 1). Using our pan-European model, our Spanish data had BAP scores of 0.905, for our Polish data was 0.960, and for our Finnish data was 0.883 while FSCT had 0.726 for our Spanish data, 0.848 for our Polish data, and 0.811 for our Finnish data. BA produced similar results, showing that our model is also a substantial improvement even without weighting for pathlength. Recall was consistently higher for our model compared to FSCT (Table 1), but precision showed more variability, with our model producing a substantially higher precision for Finland (improvement of 0.155) but lower than FSCT for Spain (a difference of -0.036) and Poland (a difference of -0.048)

a)

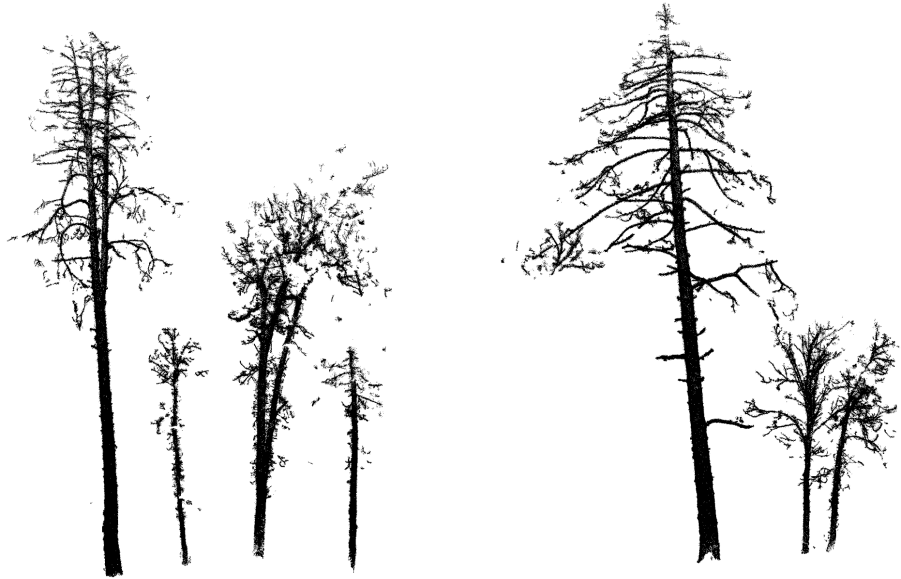

b)

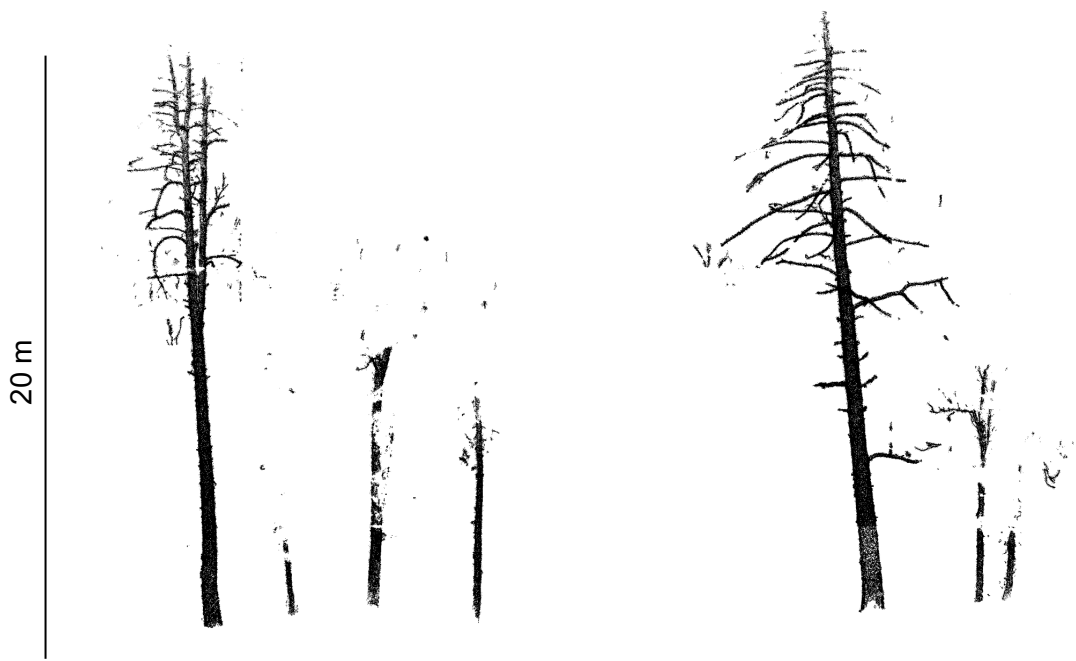

**Figure 3:** Comparison wood point output for our pan-European model (a) and FSCT (b) for two example sections data from a Spanish plot containing a mixture of *Pinus sylvestris* and *Quercus faginea*.

*Performance of our pan-European model and FSCT on others' data*

With the open datasets with and without reflectance, our model outperformed FSCT consistently with both BA (China: our model 0.912 vs FSCT 0.814; Cameroon: our model 0.952 vs FCST 0.840; Germany our model 0.929 vs FSCT 0.800; Finland our model 0.825 vs FSCT 0.793) and recall (China: our model 0.878 vs FSCT 0.646; Cameroon: our model 0.921 vs FCST 0.691; Germany our model 0.898 vs FSCT 0.614; Finland our model 0.890 vs FSCT 0.689), but for precision FSCT outperformed our model across all data (China: our model 0.679 vs FSCT 0.781; Cameroon: our model 0.921 vs FCST 0.931; Germany our model 0.895 vs FSCT 0.939; Finland our model 0.488 vs FSCT 0.630).

For BA, our model performed similarly on our own data as on the unseen third party data (0.905-0.971 vs 0.825-0.952), but results for recall were slightly better on our own data (0.821-0.954 vs 0.881-0.920). Despite being trained on data with reflectance, our model outperformed FSCT on all unseen datasets without reflectance (BA improved by 10.0% for China, 11.2% for Cameroon, 12.9% for Germany and 3.2% for Finland) and our BA on completely independent, unseen test data was 90.05%, which was comparable to the 93.7% BA achieved on our own validation data. An example of the performance of the model on a single tree from the Cameroon dataset is shown in Figure S1.

*Runtime comparison: our model vs FCST*

The improvements to semantic accuracy compared to FSCT that we demonstrate here come at no additional computation cost and runtime despite the reduced downsampling. As an example, for a point cloud of 10,743,596 million points on our hardware, our model required 46 seconds for processing, whereas the standard implementation of FSCT completed the task in approximately 86 seconds. Running FSCT at the same voxel size as ours (2 m) took 307 seconds (although we note that FSCT was not optimised for this voxel size) and visual inspection suggests that the 2 m voxel size produced visually worse FSCT outputs (see supplementary materials Figure S2). Running our model with voxels set to 6 m, consistent with the standard FSCT implementation, takes 22 seconds. However, visual assessment indicates a decrease in performance, particularly with an increase in false positives around small branches (see Supplementary Materials, Figure S3).

**Discussion**

This study presents an improved semantic segmentation model for TLS forest point clouds, compared to a widely used approach. Although also showing strong overall performance, our model focused on improvements in classification in tree crown peripheries, which are important to plant structural ecology. We found evidence for the value of multi-ecosystem and multi-biome model training, pointing towards reliable generic models that work well in multiple ecosystems. To encourage such further improvements, we make our code, model weights and labelled data open source (see 'Open Access' below).

Our model consistently outperformed the current widely used FSCT at predicting leaf and wood labels within TLS point clouds in multiple datasets and ecosystem types, including on data not collected by us, with different sensors, and both with and without reflectance information. It is, however, important to note that FSCT was designed for transferability between different sensors and data resolutions whereas our model is highly tailored to high resolution TLS point clouds and remains untested on other, lower resolution, data types, such as UAV sensors. Improvements over FSCT were significant across all ecosystem types, highlighted by both balanced accuracy and balanced accuracy weighted by pathlength. Given the significance of accurately labelling leaf and wood within point clouds for ecological inference and biomass calculations (Disney et al., 2018; Calders et al., 2020), we expect our model to have positive impact on the feasibility of building accurate wood representations (for example for carbon assessment), particularly in ecosystems where scanning during leaf off conditions is not feasible. Incorporating the presented semantic segmentation into instance segmentation pipelines e.g. TLS2trees (Wilkes et al. 2023) should improve performance by improving the labelling of leaf and wood at crown peripheries. In densely intertwined crown regions accurate instance segmentation becomes challenging and users must rely on intensive manual cleaning to label material as belonging to one tree or another (Burt et al. 2019). Improvements in classification of the smallest woody elements, and therefore improvement of the reconstruction of the woody skeleton, will reduce the size of regions of uncertain provenance in dense canopies.

Whilst showing the strongest performance of all models across all datasets for most metrics, our pan-European model showed slightly worse performance for precision for data versus FSCT across completely independent unseen data, whilst our model had higher recall than FSCT across all test data. Our model therefore classified more leaf points as wood in these datasets, which may be caused by our focal loss function forcing the model to learn from difficult points, or by other context (ecosystem-specific) features of these data. The trade-off between recall and precision can be fine-tuned by adjusting the classification threshold during inference. While thresholds can be optimised for specific applications by users, we here adopted the conventional 0.5 threshold for its simplicity and to facilitate fair comparisons with other methods. This approach provides a balanced baseline, though users can easily modify the threshold to suit particular use cases or to prioritise either precision or recall as needed.

*Ecosystem-specific vs generic models*

With deep learning models becoming increasingly large, requiring an ever-increasing degree of computational power and more and more data, the relative value of generic vs specifically trained models that are highly context specific is an open question. In this context, different sensors, ecosystems and even sampling strategies can vary to produce very different data structures which may require different models (Lines et al. 2022b). Here we directly compare the performance of biome-specific models against a model trained across multiple biomes and generally observe only very small differences in performance, and mostly improved performance by the more general model. Strong differences between performance with data from different forest types may be more likely attributed to differences in functional types (e.g., evergreen vs. deciduous broadleaf), which are mixed in our data, including within each biome. Greater discrepancies in performance between generic and tailored models may be observed between data from different sensors, where resolution and precision set constraints on classification, although we found good performance on two datasets collected with different sensors to ours, and without reflectance information. Our evidence suggests generic models can perform very well across a range of contexts and metrics, pointing towards reliable generic tools for TLS users, though we note that we focus here primarily on European forests, with at most two canopy layers, and performance on very dense multi-layer tropical forests is not tested.

*A new metric for context dependent model evaluation*

A large portion of woody points in TLS forest point clouds are concentrated within trunks and large branches where classification is more straightforward. As the woody skeleton tapers, the wood-to-leaf ratio decreases, resulting in the target class becoming smaller and more spatially mixed with the negative class, necessitating a targeted evaluation in these zones. Our framework is designed to assess TLS processing for ecologically important information, and so we devised a new accuracy metric that aligns with the needs of this application, specifically by weighting evaluation within crown and at their peripheries, where leaf points dominate. In these regions, woody points constitute only a proportion of the total points, and thus their accurate classification has limited impact on overall tree or forest statistics, but accurate classification is key to separating tree crowns well. We suggest that DL models in ecological applications, customised to specific contexts, should consider similar approaches to enhance clarity for end-users, rather than simply presenting standard DL metrics alone. For example, ecologists concerned with plant hydraulics, growth, and spatial occupation may require precise woody skeletons at high path lengths and so value model assessment with BAP, whereas for biomass assessment, the significance of these regions is likely less because the majority of aboveground biomass is in trunks and large branches, and BA may be sufficient.

*Future model development directions*

High-quality benchmarking datasets are essential for accurately assessing model performance on complex 3D data. Semantic labelling of leaf and wood structures in 3D is a challenging and time-consuming task, further compounded by the limitations of less powerful instruments and the complexity of ecological contexts, such as tall, layered canopies. Better datasets will lead the way towards more effective method development, and we found that not all open datasets stood up to visual inspection. While our retrained FSCT model showed improved performance on our validation data compared to its off-the-shelf variant, it exhibited decreased performance on completely independent unseen data. This performance discrepancy may be attributed to the significant differences in tree morphology between our dataset and the unseen datasets. In contrast, our model performed similarly on our and the third party datasets. The tropical trees in the unseen data are approximately 50 m tall with crowns spanning 25 m, contrasting with our smaller stature trees that average 25 m in height with significantly smaller crowns (approximately 10m). Our model's superior transferability between these different scales is likely due to the neighbourhood rescaling function (Qian et al., 2022) implemented within our PointNet++ architecture.

We found value in including reflectance information within our model, but this is rarely supplied in open datasets. One reason for its perceived lack of value and low reliability may be its noisiness and sensitivity to data collection conditions. To address these issues, we implemented a novel gating reflectance mechanism. This feature allows our model to flexibly adapt to the quality and availability of reflectance data. When reflectance information is informative, the model uses it; when it is noisy or unreliable, the gating mechanism reduces its influence. This approach prevents the model from overly depending on reflectance data, ensuring robust performance across varying data conditions. Our gating mechanism thus enhances the model's versatility, allowing it to maintain high performance whether reflectance data is present, absent, or of varying quality. We encourage researchers to publish reflectance alongside positional xyz data and in a format as close to raw as possible to enable bespoke downsampling procedures to be developed and tested.

Our model was trained solely on European forest datasets, and while our data is composed of varied forest types, our training is still constrained by data availability. Ultimately, incorporating other sensors with reflectance/intensity into training would likely strengthen our models' capabilities further. A key application of TLS is the improvement of carbon accounting in the wet tropics (Disney, 2019) where trees are larger than those in our data, have more variety of leaf size, and are evergreen. Such ecosystems pose significant additional challenges for classification at tree crown peripheries but without more well labelled data from a variety of ecosystems the performance of our - and others' - models for this task cannot be evaluated. Finally, more tests on lightweight but context-focused models could potentially make models more accessible to those without large computational resources but working within a specific ecological context.

*Conclusions*

Our study showcases the efficacy of our new model for semantic segmentation of wood and leaf in TLS point clouds across various European biomes and unseen ecosystems. Using a newly developed metric we show that, compared to the most widely adopted existing approach, our model excelled in identifying small features such as twigs and branches in the outer canopy, critical for a variety of ecological applications. By testing against meticulously labelled data, including in the challenging upper canopy regions, we demonstrated improved performance of our model. Leveraging reflectance and strategic downsampling techniques, our model preserves small woody features while driving efficient sampling throughout the extraction layers. Furthermore, we provide comparative insights between biome-specific and pan-European models, underscoring the versatility and robustness of our approach. Lastly, we contribute to open data and open-source software by sharing our data, code and model weights, facilitating further advancements in the field.

**Open Access**

*Data used in this study are available at:*

UNF:6:9U7BGTgjjsWd1GduT1qXjA== [fileUNF]. Distributed under a Creative Commons Attribution 4.0 International Deed.

## Data Availability

Plot-level semantically labelled terrestrial laser scanning point clouds (1.0) [Data set]. Zenodo. https://doi.org/10.5281/zenodo.13268500

*The model code and weights are available at:*

Owen, H.J. (2024) *PointsToWood*. Available at: https://github.com/harryjfowen/PointsToWood/ (Accessed: 18 October 2024). *The software presented here can be run locally or on high performance computers.

## Acknowledgements

H.J.F.O., S.W.D.G. and E.R.L. were funded by a UKRI Future Leaders Fellowship awarded to E.R.L. (MR/T019832/1). M.J.A. was supported by the UKRI Centre for Doctoral Training in Application of Artificial Intelligence to the study of Environmental Risks (EP/S022961/1).

## Author Contributions

*Harry Owen and Emily Lines conceived the ideas and designed methodology; Harry Owen, Emily Lines and Stuart Grieve collected the data. Harry Owen analysed the data and led the writing of the manuscript. All authors contributed critically to the drafts and gave final approval for publication.*

**Supplementary:**

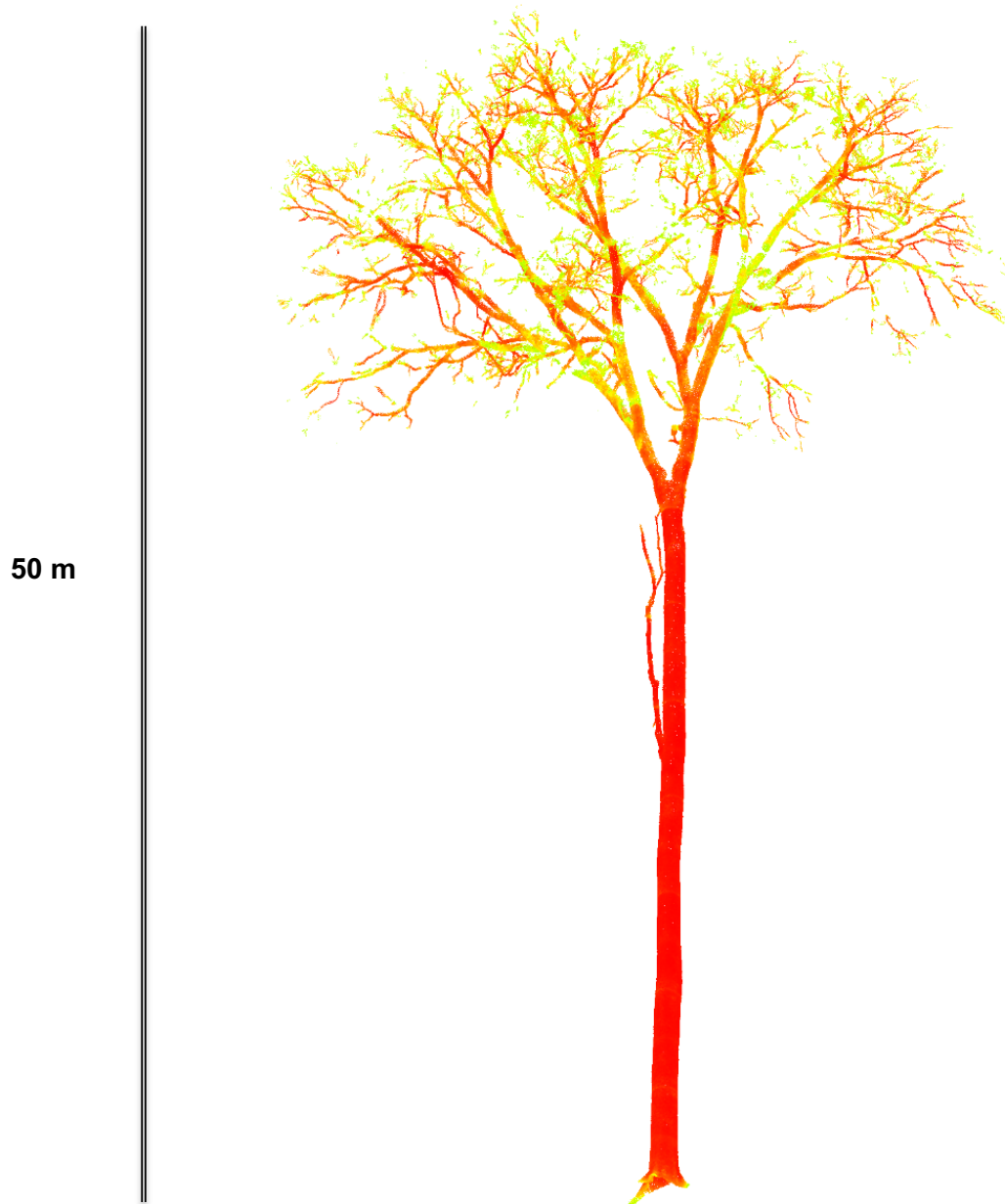

**Figure S1:** Unseen tropical labelled data from the test dataset from Eastern Cameroon Di Wang et al., 2022 without reflectance passed through our pan-European model. Yellow through to red initiative of probability of wood, here thresholded to above 0.55 for visual purposes.

a)

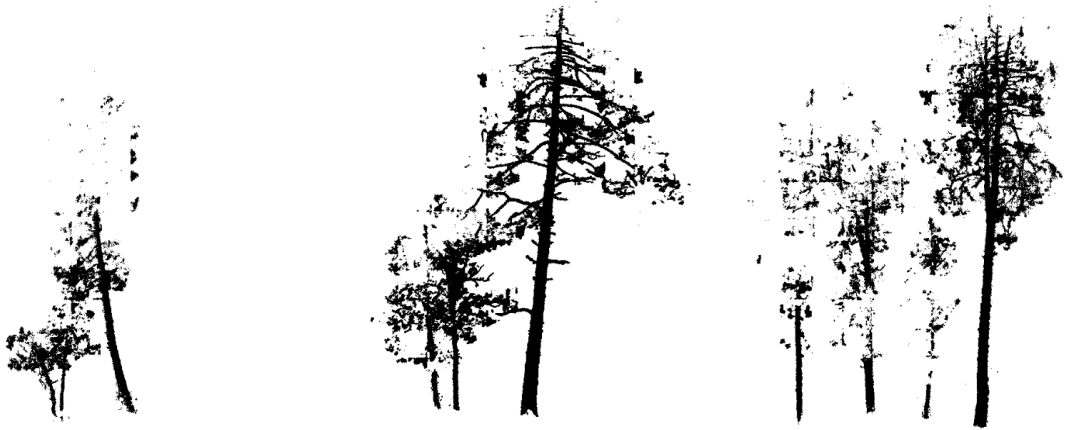

b)

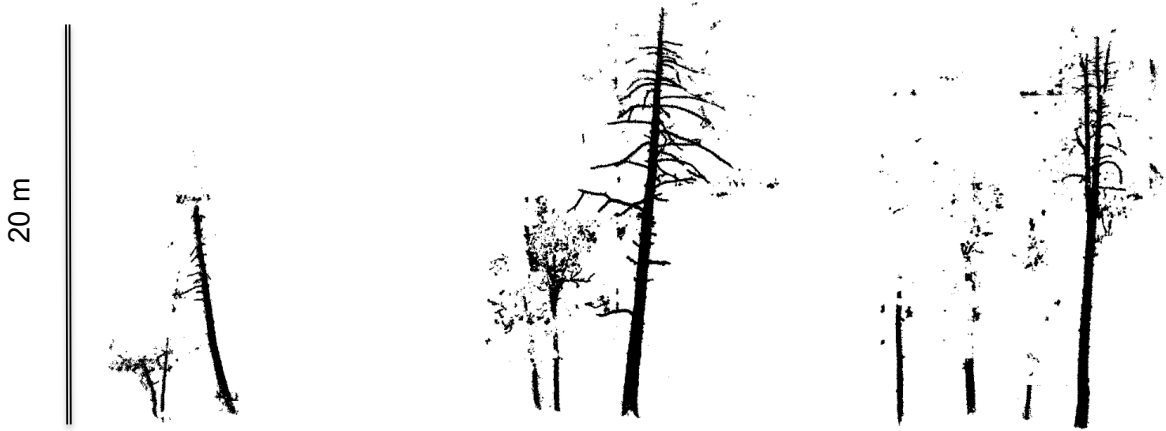

**Figure S2:** FSCT output with a) 2 m voxel input resolution and b) 6 m resolution on our Spanish dataset

a)

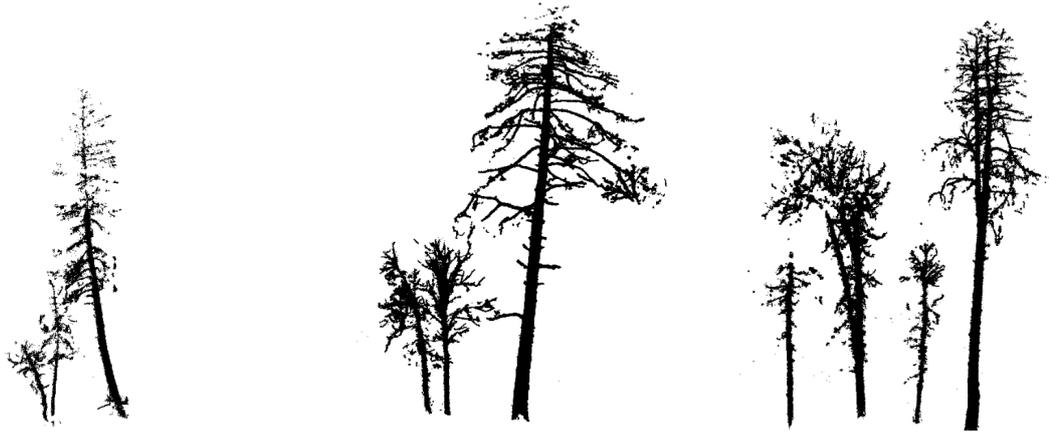

b)

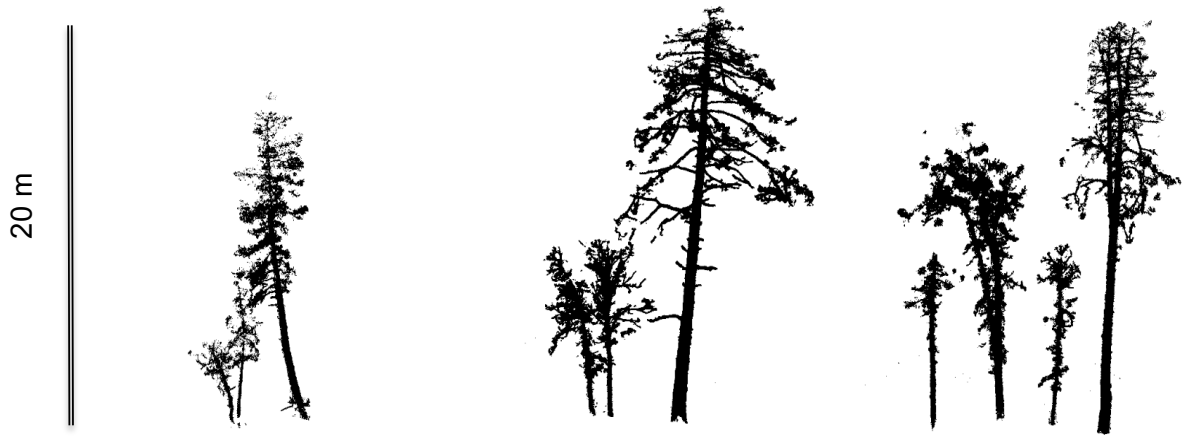

**Figure S3:** Our pan-European model output with a) 2 m resolution and b) 6 m resolution on our Spanish dataset

**Table S1:** Comparison of our models to the off-the-shelf FSCT model, with no re-training at all.

| *Pan European model* vs FSCT (off-the-shelf version) on our data | | | | |
|---|---|---|---|---|
| *Country* | *Balanced accuracy (BA)* | *Precision* | *Recall* | *Balanced accuracy weighted by pathlength (BAP)* |
| *Our data (TOF with reflectance)* | | | | |
| Finland | **.905**, .837 | **.945**, .656 | **.821**, .771 | **.885**, .798 |
| Poland | **.971**, .841 | **.864**, .847 | **.954**, .692 | **.960**, .786 |
| Spain | **.934**, .769 | **.909**, .707 | **.906**, .595 | **.906**, .726 |
| *Third party data* | | | | |
| Cameroon (TOF no reflectance) | **.952**, .866 | **.921**, .807 | **.920**, .770 | n/a |
| China (TOF no reflectance) | **.914**, .837 | **.679**, .480 | **.881**, .762 | n/a |
| Germany (TOF w/reflectance) | **.929**, .814 | **.895**, .861 | **.897**, .672 | n/a |
| Finland (phase-shift w/intensity) | **.825**, .812 | **.488**, .719 | **.890**, .694 | n/a |